\documentclass[10pt,twocolumn,a4paper]{esaAI}

\usepackage{comment}
\usepackage{tikz,lipsum,lmodern}
\usepackage[most]{tcolorbox}
\title{AI Assistants for Spaceflight Procedures: Combining Generative Pre-Trained Transformer and Retrieval-Augmented Generation on Knowledge Graphs With Augmented Reality Cues}

\author[1,3]{Oliver Bensch\thanks{Corresponding author.}}
\author[1,2]{Leonie Bensch}
\author[2]{Tommy Nilsson}
\author[2]{Florian Saling}
\author[2]{Bernd Bewer}
\author[1]{Sophie Jentzsch}
\author[1]{Tobias Hecking}
\author[4]{J. Nathan Kutz}

\affil[1]{German Aerospace Center, Institute for Software Technology}
\affil[2]{European Space Agency, European Astronaut Center (ESA-EAC)}
\affil[3]{AI Institute in Dynamic Systems, University of Washington, Seattle, WA USA}
\affil[4]{Department of Applied Mathematics, University of Washington, Seattle, WA USA}

\begin{document}

\makeCustomtitle

\begin{abstract}
This paper describes the capabilities and potential of the intelligent personal assistant (IPA) CORE (Checklist Organizer for Research and Exploration), designed to support astronauts during procedures onboard the International Space Station (ISS), the Lunar Gateway station, and beyond. We reflect on the importance of a reliable and flexible assistant capable of offline operation and highlight the usefulness of audiovisual interaction using augmented reality elements to intuitively display checklist information. We argue that current approaches to the design of IPAs in space operations fall short of meeting these criteria. Therefore, we propose CORE as an assistant that combines Knowledge Graphs (KGs), Retrieval-Augmented Generation (RAG) for a Generative Pre-Trained Transformer (GPT), and Augmented Reality (AR) elements to ensure an intuitive understanding of procedure steps, reliability, offline availability, and flexibility in terms of response style and procedure updates.
\end{abstract}

\section{Introduction}

Amidst a surging interest in human space exploration, the NASA-led Artemis program is seeking to establish a sustained human presence on the Moon \cite{smith2020artemis}.
These efforts are targeting the Moon’s south pole region, where strategically important resources, such as water ice, are believed to be preserved at the bottom of permanently shadowed craters. Yet, with the lunar libration causing the polar region to move out of the Earth’s view periodically, recurring communication blackouts will be unavoidable, limiting support from Earth \cite{weber2021artemis,horneck2003humex,braly_augmented_2019,haney2020apollo,landgraf_lunar_2021}.  
Hence, a small space station, referred to as \textit{Lunar Gateway}, will be constructed in orbit around the Moon. The Lunar Gateway should facilitate crewed missions on the lunar surface by functioning as a communication relay to Earth. Moreover, the station will feature docking ports for spacecraft and be able to accommodate several astronauts, enabling the monitoring of lunar surface activities while also facilitating research experiments in lunar orbit \cite{johnson2021gateway}. 

To ensure the efficiency and safety of operations, Gateway will incorporate standardized procedures and protocols, leveraging decades of operational experience from the International Space Station (ISS).

These standardized procedures generally include workflow guidance for assembling, servicing, and maintaining of equipment, scientific experiments, and emergency procedures among others.

Yet historically, astronauts performing procedures in space (e.g., onboard ISS or during the Apollo missions) relied on relatively rudimentary tools, such as manual checklists, which often left them dependent on continuous supervision from mission control centers on Earth, which will not be possible during future missions to the Moon and beyond \cite{apollo11_1969, mindell2011digital,hersch2009checklist, marshburn2003independent, kintz2016communication}. 

Therefore, to ease the work of astronauts during procedure completion, intelligent personal assistants (IPA) are attracting growing interest in the human spaceflight domain for their ability to facilitate natural, hands-free, and unobtrusive support for procedural tasks, thereby reducing the astronaut's need for direct supervision from Earth, through guiding the astronaut through a given procedure.

In this regard, a key milestone was the 2018 ISS deployment of CIMON, a Watson-powered IPA  \cite{DLR2018CIMON}. Its successor, the CIMON-2, launched in 2019 and included enhanced features, such as visual recognition and emotion analysis, but required internet connection and lacked autonomous learning \cite{schmitz2020towards}. Similarly, in 2022, Amazon's Alexa was integrated into the Artemis I lunar mission \cite{AlexaArtemisI}. Yet both IPAs are limited by their rule-based nature, meaning that they can only respond to a pre-defined set of questions regarding procedure steps. Meanwhile, the AI4U assistant, developed using a reinforcement learning approach, is being tested under simulated Mars conditions during a Mars analog study with the goal of eventually being deployed and further assessed aboard the ISS \cite{shashkovaa2022study, navarro2023eclss, spoon_ai4u}. Unlike rule-based IPAs like CIMON and Alexa, AI4U learns new skills during operation but requires extensive training sessions.

Consequently, contemporary IPAs designed to support astronauts suffer from important limitations, including the lack of offline availability, limited flexibility regarding their range of answers, and delays in updating procedures (due to their rule-based nature or the requirement of extensive training). 

Moreover, these IPAs provide relevant procedure information primarily in the form of text-based and/or spoken instructions. Although they sometimes make use of visual aids, including pictures and videos, they are currently incapable of delivering more complex and spatially intuitive visual information, such as 3D representations of procedure steps or elements. 

In the past, this increased the workload for astronauts, as they had to switch their gaze between the checklist and the current task, while requiring complex mental rotations to interpret visual information from 2D pictures or videos, causing cognitive dissonance between procedure instructions and the real task \cite{braly_augmented_2019}.

To address these shortcomings, we propose a novel IPA called \textit{CORE}, specifically designed to support astronauts during upcoming human spaceflight missions. CORE is designed to be modular, allowing for the integration of open-source and offline deployable components. These components include Generative Pre-Trained Transformers (GPT), Retrieval-Augmented Generation (RAG), Knowledge Graphs (KG), speech recognition, and speech synthesis. This unique combination enables CORE to effectively emulate human conversation, while the astronaut is visually supported using augmented reality (AR) elements displayed in a head-up display (HUD).

\begin{figure}[t]
    \centering
    \includegraphics[width=.95\columnwidth]{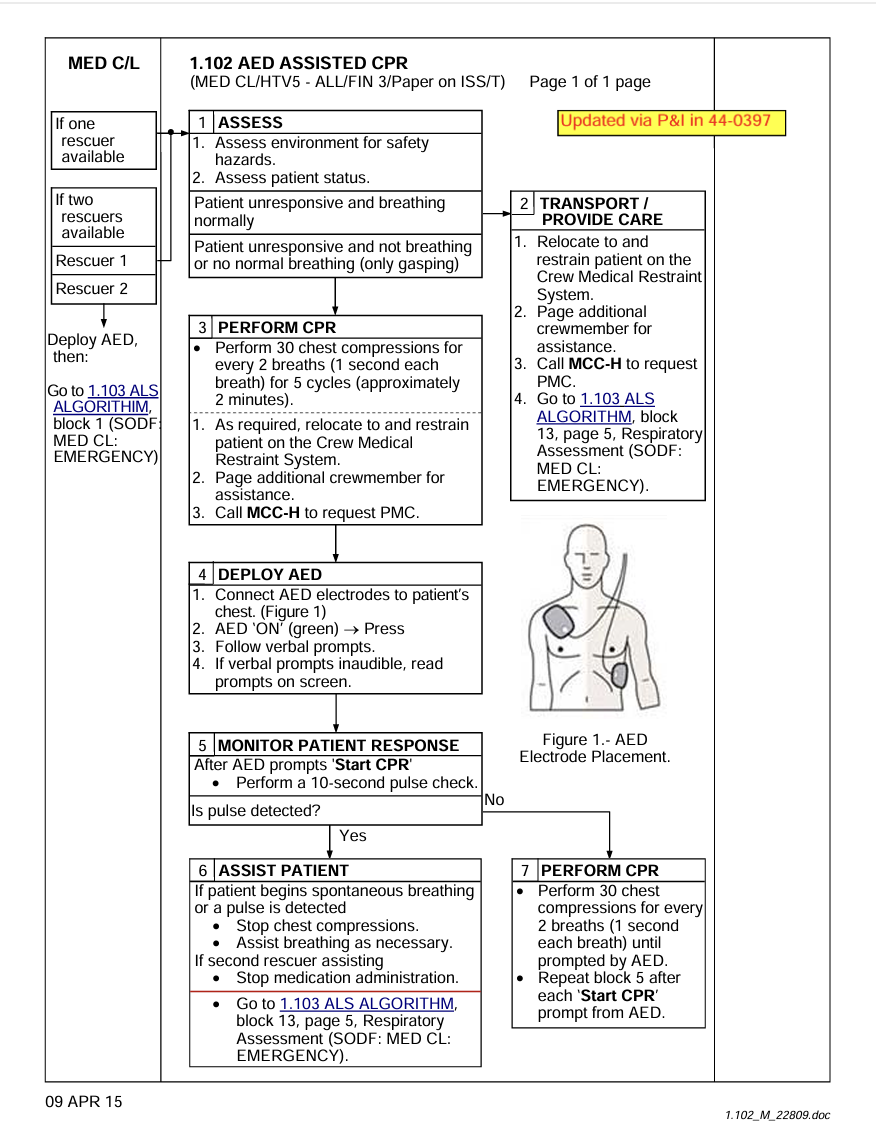}
	\caption{Example ISS procedure containing specific steps for a CPR, featuring pictures and location-depended information about the placement of electrodes on the body of the patient \cite{NASA2016ISS}. 
	}
	\label{fig:CPR}
\end{figure}

\section{IPAs for Future Space Missions}

In the following sections, we elaborate four key criteria that should be met by a future IPA for astronaut support: (1) reliability, (2) flexibility, (3) offline availability, and (4) the integration of visual 3D information. We furthermore describe how these criteria can be fulfilled using state-of-the-art technology. Drawing on this analysis, we present our IPA prototype \textit{CORE}, tailored to assist astronauts during procedure completion. We conclude by presenting an overview of our planned future work.

\noindent {\bf Reliability:} 
Reliability is a critical requirement for IPAs supporting astronauts during procedure completion, as it directly affects crew safety and the success of the mission. Firstly, any misinformation or errors in the system regarding procedure steps could potentially lead to fatal consequences. Secondly, unreliability in conversational agents often leads to user disengagement, distrust, and frustration \cite{tulshan2019survey}. Hence, it can be assumed that the slightest unreliability during procedure support would lead to the disengagement of the astronauts and a preference for the usage of traditional tools. 

To 
significantly improve the reliability of astronautical procedure support, we propose the integration of GPTs with advanced data retrieval technologies. 

GPT models, recognized for their ability to generate human-like responses without explicit instructions, play an important role in modern AI systems \cite{bubeck2023GPT4GeneralKnowledge}. However, their reliance on training datasets poses challenges, as the information generated may be inaccurate or misleading, especially if it falls outside the scope of their training data \cite{WhyChatGPTHallucinations}. While model fine-tuning can adapt GPTs to specific domains by incorporating domain-specific knowledge, this process is computationally intensive and lacks transparency, as it provides no sources for the generated responses \cite{WhyChatGPTHallucinations}.

To address these limitations and enhance reliability, we propose the integration of GPT models with RAG and KGs. This approach leverages the generative capabilities of GPTs alongside a robust retrieval system that utilizes a structured knowledge database \cite{he2024gretriever}. By interlinking documents, metadata, and multimodal data within a KG, connected by references, topics, keywords, or related company or person information, such a system ensures that all information retrieved and utilized by the AI is accurate and up-to-date. User queries initiate a vector search within this database to identify relevant information, which is added to the GPT query \cite{lewis2020RAGPaper}. This approach not only reduces the likelihood of erroneous outputs (hallucinations) but also improves the system's trustworthiness by allowing astronauts to verify the answer through the linked information used for its generation. Additionally, we suggest the enhancement of system transparency and user trust through the Graph of Thoughts (GoT) paradigm, which can be used to visualize the reasoning process of AI, enabling users to interact with and adjust the AI's thought process in real-time \cite{besta2023GoT, WuAiChains2022CHI}.

Furthermore, to further improve the astronauts understanding of the systems capabilities, we propose displaying confidence scores next to retrieved data during RAG, indicating the reliability of the information retrieved \cite{rechkemmer2022confidence, zhang2020effect}. This method, complemented by using KGs linked with established ontologies like schema.org \cite{guha2016schema}, ensures astronauts have access to rich, well-organized, and easily navigable information, crucial for decision-making in mission-critical scenarios \cite{Auer2007DBPedia, Vrande2012Wikidata, Zhu_2022MMKG}.

\noindent {\bf Flexibility:}
Importantly, the assistant's adaptability to new situations and requirements, ranging from routine procedures to unexpected emergencies, is a critical factor. Furthermore, the assistant should be flexible enough to accommodate the varied personal styles of astronauts during procedure completion. For instance, one astronaut might prefer interacting with the assistant in a voice-based manner, while another may opt for displaying written procedure steps. Additionally, preferences in interaction styles may differ among astronauts, as one may prefer an assistant that explains just the necessary steps, while another might favor an assistant with its "own character" that can engage in casual conversations, similar to a colleague. Lastly, one crucial aspect for such a system is that procedures can be flexibly adapted due to updates in the procedures themselves.

Regarding the need for flexible response styles, such as shifting between casual and formal tones, GPT-based assistants exhibit a level of versatility, unlike classic IPAs limited to predefined responses. GPT models can be steered continuously depending on requirements and preferences using steering vectors \cite{konen2024style}. The latest GPT models are capable of document writing and editing and are evaluated to modify, add, or remove components of Knowledge Graphs or document databases through natural language conversations \cite{10387715}. Furthermore, RAG enables such a system to access the latest updated information directly after indexing into a vector database, without the need for retraining of the GPT model \cite{lewis2020RAGPaper}. Linked information in a mission KG can be added or edited in a natural conversation without changing the approved procedure's content.

\noindent {\bf Offline Availability:}
Given the risk of radio blackouts and communication delays inherent in human spaceflight (and lunar south pole operations in particular), one additional crucial aspect is the need for offline availability when developing future IPA concepts to support astronauts during procedure completion.

While KG databases are widely available for offline deployment, leading GPT models like Gemini 1.5 or GPT-4 and leading speech synthesis modules are only available online \cite{geminiteam2024gemini, openai2024gpt4, WhisperPaper} due to their extensive computational requirements and intellectual property protection. However, published open-source GPT models like Mistral's Mixtral 8x7b \cite{touvron2023llama} are narrowing the performance gap while also reducing model size \cite{jiang2024mixtral} and, consequently, hardware requirements. Some models, like Meta's LLaMA 3.1 405B model \cite{llama2024herd}, are even achieving a similar performance. These models can benefit from further optimization methods such as low-rank adaptation (LoRa) \cite{hu2021lora} and low-bit quantization \cite{zhao2023atomQuantization}, which enable small models like Microsoft's Phi-3 to run on mobile devices without the requirement of a dedicated GPU \cite{hu2021lora, carreira2023GPTMobile, abdin2024phi3}. This approach allows for a streamlined and efficient AI assistant setup in astronaut suits for deep space missions, where critical functionalities like speech recognition are embedded in the suit, while more computationally intensive tasks are handled by the larger GPT model and database located at a nearby base, connected through reliable and high-speed radio communication links.

Overall, we propose that an offline IPA based on a GPT model with an RAG architecture querying a KG for astronauts with low hardware requirements is already feasible. The quality of answers generated by open-source GPT models like Llama 3.1 can be compared to GPT-4 \cite{llama2024herd}. Although, the quality of voices generated by open-source Text-To-Speech models has yet to match that of online available models like OpenAI's \cite{OpenAITTS}.

\noindent {\bf Combining Voice, Text, and Visual 3D Information:}
Text and picture-based procedures pose the risk of significantly increasing the workload of astronauts due to 1) the need to switch attention back and forth between the visual spaces of the task and the procedure instructions and 2) the mental effort of mapping instruction steps to the real equipment \cite{braly_augmented_2019, yeh1998effects}. 

To address these issues, past research has demonstrated that HUD AR technology can ease the work of astronauts during procedural tasks onboard the ISS by superimposing computer-generated instructions directly onto real equipment \cite{braly_augmented_2019, markov2013pilot, helin2018user}. Additionally, the combination of AR and voice commands offers hands-free interaction without requiring astronauts to divert their gaze to another display \cite{cardenas2021reducing, rometsch2022design}. Moreover, recent research has shown that combining LLMs and AR HUDs is beneficial for the completion of procedural maintenance tasks \cite{xu2024augmented}. Therefore, we suggest combining the advantages of AR technology and LLMs to allow for intuitive interaction between astronauts and procedure elements. We furthermore propose that graphically displaying relational information from the KG could further aid astronauts in understanding linked concepts, components, documents, or images during critical situations, enhancing the assistant's explainability \cite{nararatwong2020knowledge, serrano2022vowlexplain}.

\section{Results - CORE Prototype}
In our initial experiments, we indexed ten publicly available procedures \cite{NASA2016ISS} into a vector database and combined the documents with metadata such as the edit date and related images in the \textit{ArangoDB} graph database. Subsequently, we implemented a RAG system based on the GPT-4 model and indexed textual information into the Elasticsearch vector database to achieve the best possible results during our experiments, while keeping the CORE system modular for easy replacement with newly released offline models, such as the Llama 3.1 405B model, which offer similar performance \cite{llama2024herd}. A speech user query is first converted into text using \textit{Whisper} \cite{WhisperPaper}. Relevant procedure information is retrieved from the vector database and added to the user query, along with linked information in the graph, using the following prompt template:

\begin{quote}
  You will be presented with a matching procedure enclosed by three quotation marks ('''). If a question is asked, respond with either the next step or only the first step. Specify the relevant step and repeat the text of the procedure step verbatim. If the information does not correspond to the question or if information is missing, state that this is the case.
  '''\textit{[Procedure] [Graph information]} 
  '''
\end{quote}

Furthermore, using the following system-prompt, we directed the GPT-model to respond exclusively to procedure-related questions and return the current step as well as information about linked images in a separate format to be displayed to the user via the interface if necessary: 

\begin{quote}
  You are a helpful assistant for astronauts, answering questions about provided procedures. If asked a question, respond with either the next step or the first step only. Name the corresponding step as <<STEP [NUMBER]>> and repeat the text of the procedure step word for word. If a figure or other data is referenced, include <<SHOW FIGURE [NUMBER]>> in your answer.
\end{quote}

Our first tests have shown that depending on the system-prompt, the GPT-4 model is able to interpret the given procedures and answer related user questions correctly. 
For example procedure steps were returned verbatim for the user question:

\begin{quote}
Hi, I have a person that is not breathing. I have already requested PMC. What was the fourth step of the ISS CPR procedure?
\end{quote} 

The system returns the corresponding step word by word and adds additional information to correctly display the step and the linked image.

\begin{quote}
<<STEP 4>> - DEPLOY AED:
Connect AED electrodes to patient's chest. (See Figure 1)
AED ON (green) → Press
Follow verbal prompts.
If verbal prompts inaudible, read prompts on screen.
Continue with "Step 5" <<SHOW FIGURE 1>>
\end{quote}

When the system is asked for graph data that is not added as a full sentence, such as "Last update: 09 April 2015" the system formulates a full sentence. For instance, if the user asks:

\begin{quote}
When was the procedure last updated?
\end{quote} 

the system replies:

\begin{quote}
The CPR procedure on the ISS was last updated on 09 April 2015.
\end{quote} 

Moreover, unrelated questions regarding other topics than procedure-related information are not answered by CORE. 

\section{Discussion}
Our work highlights the significance of IPA technologies for future ISS, Gateway missions, and missions beyond. We propose a solution based on state-of-the-art natural language processing methods. Our primary focus is to develop a versatile IPA that seamlessly adapts to a range of tasks and interaction styles through the use of GPT models. These models surpass current IPAs, being capable of handling unforeseen situations with enhanced reasoning, creativity, and task-solving abilities. Reliability is another crucial factor, as it directly impacts astronauts' trust and engagement with the system. We recommend the integration of RAG and KG technology with GPT models, ensuring accurate information retrieval and enhancing trustworthiness, which is vital for high-stakes space operations. Additionally, we emphasize the need for offline availability of these systems, considering the communication limitations in space. Recent technological advancements suggest the feasibility of efficient, offline IPAs that require minimal hardware resources. Finally, we propose a combination of voice-based and visual information, utilizing AR cues to enhance the intuitive understanding of complex spatial information such as checklists, navigation, or KG relations. 

The second version of CORE runs completely offline and is based on Llama 3.1 and integrates AR HUD features into our prototype. We will test it in a user study at the European Astronaut Centre (ESA EAC) to assess user experience and performance metrics.

\printbibliography
\addcontentsline{toc}{section}{References}

\end{document}